\pdfoutput=1
\documentclass[11pt]{article}

\usepackage[]{EMNLP2023}

\usepackage{times}
\usepackage{latexsym}

\usepackage[T1]{fontenc}
\usepackage[utf8]{inputenc}

\usepackage{microtype}

\usepackage{inconsolata}
\usepackage{enumitem}
\usepackage{multirow}
\usepackage{booktabs}
\usepackage{arydshln}
\usepackage{graphicx}
\usepackage{CJKutf8}
\usepackage{makecell}
\usepackage{amsmath}
\usepackage{amssymb}
\usepackage{multicol}
\usepackage{booktabs,colortbl}

\usepackage{inconsolata}
\usepackage{enumitem}
\usepackage{booktabs,colortbl}
\usepackage{graphicx}
\usepackage{url}
\usepackage{CJK}
\usepackage{multirow}
\usepackage{multicol}

\usepackage{arydshln}
\usepackage{float}
\usepackage{caption}
\usepackage{amsmath}
\usepackage{amssymb}
\usepackage{bbold}
\usepackage{tabularx}

\definecolor{forestgreen}{rgb}{0.13, 0.55, 0.13}

\title{Findings of the WMT 2024 Shared Task on Discourse-Level Literary Translation}

\author{Longyue Wang, Siyou Liu, Chenyang Lyu, Wenxiang Jiao,  Xing Wang, Jiahao Xu,\\ \bf Zhaopeng Tu, Yan Gu, Weiyu Chen, Minghao Wu, Liting Zhou,\\ \bf Philipp Koehn, Andy Way, Yulin Yuan\\ vincentwang0229@gmail.com}

\begin{document}
\maketitle
\begin{abstract}

Following last year, we have continued to host the WMT translation shared task this year, the second edition of the {\em Discourse-Level Literary Translation}. We focus on three language directions: Chinese$\rightarrow$English, Chinese$\rightarrow$German, and Chinese$\rightarrow$Russian, with the latter two ones newly added. This year, we totally received 10 submissions from 5 academia and industry teams. We employ both automatic and human evaluations to measure the performance of the submitted systems. The official ranking of the systems is based on the overall human judgments. We release data, system outputs, and leaderboard at 
\url{https://www2.statmt.org/wmt24/literary-translation-task.html}.
\end{abstract}

\section{Introduction}

The organization of the WMT shared task ({\em Discourse-Level Literary Translation}) aims to explore the potential of machine translation (MT) and large language model (LLM) in overcoming the unique challenges of literary translation.
In recent years, several studies have explored the capabilities of LLMs in the fields of literary translation \cite{an2023max,lopez2023make,zhao2023dutnlp,xie2023hw,zhu2023tjunlp} and discourse-level translation \cite{wang2023document,wang2024benchmarking,wu2024perhaps,wang2024benchmarking}.

Following last year's WMT shared task~\cite{wang2023findings}, we have continued to host the second edition of the {\em Discourse-Level Literary Translation}.
Notably, we expand the datasets to include Chinese-German and Chinese-Russian (only document-level), in addition to the existing Chinese-English  (both document- and sentence-level).\footnote{\url{https://github.com/longyuewangdcu/GuoFeng-Webnovel}.} The introduction of these new language pairs presents unique challenges, particularly given the document-level nature of the data and the absence of sentence-level alignment in the new datasets.

Besides, we provide different types pretrained models and LLMs such as Chinese-Llama-2\footnote{\url{https://github.com/longyuewangdcu/Chinese-Llama-2}.} and in-domain RoBERTa and mBART \cite{wang2023disco}. Apart from automatic evaluation, we employ two human evaluation criteria: 1) {\em general quality}, covering aspects such as fluency and adequacy; 2) {\em discourse-aware quality}, including factors such as consistency, word choice, and anaphora.\footnote{Due to time constraints, we were unable to conduct A/B testing with web fiction readers this year.} Since there were only two participating teams for Chinese$\rightarrow$German, and Chinese$\rightarrow$Russian, we only use automatic evaluation. The task has Constrained Track and Unconstrained Track, however, most teams choose Unconstrained Track.

In this year, we totally received
10 submissions from 5 academia and industry
teams. We found that (1) most of the participating teams' models (after literary-domain enhancements) outperform Baseline systems (Llama-MT, Google Translate and GPT-4) in terms of d-BLEU; (2) for a certain system, there is a significant gap between the conclusions drawn from human and automatic evaluations; (3) the performance of the only Constrained Track system was close to that of the best system in the Unconstrained Track in terms of both automatic and human evaluation.

\begin{figure*}
\begin{minipage}[b]{0.5\linewidth}
    \begin{tabular}{l rrrrr}
    \toprule
    \textbf{\color{blue}Zh-En} & \textbf{\#Book} & \textbf{\#Chap.} & \textbf{\#Sent.} & \bf \#Word & \textbf{|D|} \\
    \midrule
    Train & 179 & 22.6K & 1.9M & 32.0M & 1.4K\\ \midrule
    Valid$_1$	& 22	&22	&755 & 18.3K & 832\\
    Test$_1$	& 26	&22	&697 & 19.5K & 884\\
    \cdashline{1-6}\noalign{\vskip 0.5ex}
    Valid$_2$	& 10	&10	&853 & 16.0K & 1.6K\\ 
    Test$_2$	& 12	&12	&917 & 16.7K & 1.4K\\ \midrule
    Test$_{final}$	& 12	&239	& 16.7K & 337.0K & $^*$28.1K\\
    \bottomrule
    \end{tabular}
\end{minipage}
\hfill
\begin{minipage}{0.33\textwidth}
  \centering
  \includegraphics[width=1\linewidth]{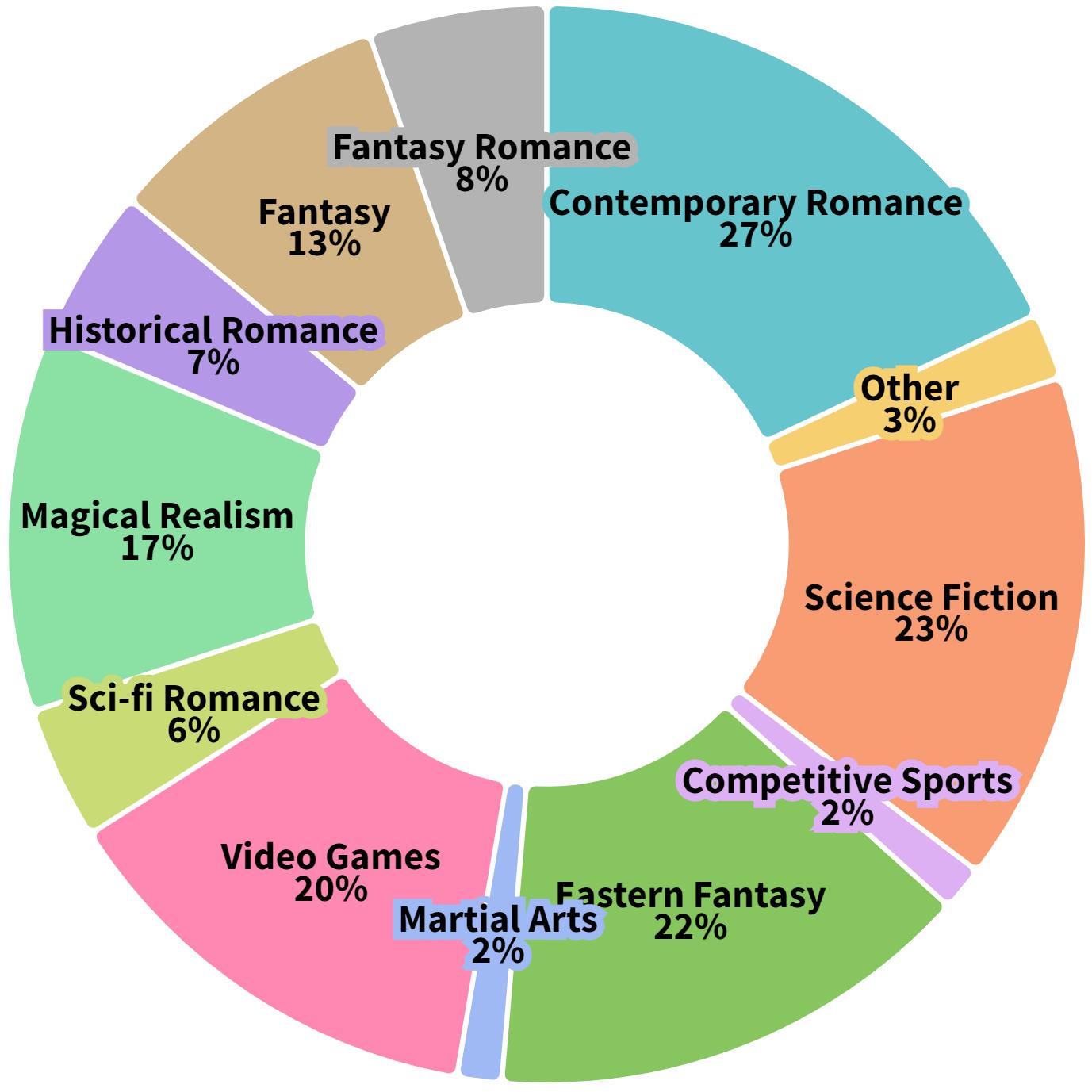}
\end{minipage}
\hfill
\begin{minipage}[b]{0.5\linewidth}
    \begin{tabular}{lrrr c lrrr}
    \toprule
    \textbf{\color{red}Zh-De} & \textbf{\#Book} & \textbf{\#Chap.} & \bf \#Word && \textbf{\color{forestgreen}Zh-Ru} & \textbf{\#Book} & \textbf{\#Chap.} & \bf \#Word\\
    \midrule
    Train & 118 & 19.1K & 25.6M && Train & 122 & 20.0K & 23.5M\\
    Valid & 11 & 11 & 16.7K && Valid & 11 & 11 & 14.5K \\
    Test & 13 & 13 & 18.8K && Test & 122 & 13 & 15.6K\\
    \bottomrule
    \end{tabular}
\end{minipage}
    \caption{\label{tab:data_statics}Data statistics of the GuoFeng Webnovel Corpus V2 in Chinese$\rightarrow$English (Zh-En), Chinese$\rightarrow$German (Zh-De), Chinese$\rightarrow$Russian (Zh-Ru) and genre distribution (Chinese$\rightarrow$English training set). The \#Book, \#Chap. \#Sent. mean the number in terms of books, chapters and sentences. The \#Word is based on target language sides. The document length (|D|) is calculated by dividing \#Word by the number of documents. In Zh-De and Zh-Ru, the |D| equals to \#Chap.}
\end{figure*}

\section{The GuoFeng Webnovel Corpus V2}
\label{sec:2}

\subsection{Copyright and License}

Copyright is a crucial consideration when it comes to releasing literary texts. Tencent AI Lab and China Literature Ltd. hold the copyright for this dataset. To support research advancement in this field, we offer the data to the research community under specific terms and conditions:
\begin{itemize}[leftmargin=*,topsep=0.1em,itemsep=0.1em,parsep=0.1em]
    \item After registration, WMT participants can utilize the corpus for non-commercial research and must adhere to the principle of fair use (CC-BY).
    \item Modification or redistribution of the dataset is strictly prohibited.
    \item Proper citation of this paper and the original download link is required.
    \item By using this dataset, you agree to the terms and conditions outlined above. We take copyright infringement very seriously and will take legal action against any unauthorized use of our data.
\end{itemize}

\subsection{New Datasets}
\label{sec:2-1}

The data processing follows the same framework as last year \cite{wang2023findings}. The data statistics of the corpus are detailed in Figure~\ref{tab:data_statics}. 
\begin{itemize}[leftmargin=*,topsep=0.1em,itemsep=0.1em,parsep=0.1em]
\item {\bf Zh-En set}: This is same as the GuoFeng Webnovel Corpus V1. The \emph{training set} comprises 22,567 continuous chapters from 179 web novels, spanning 14 genres such as fantasy and romance. To enable self-evaluation of model performance, we offer two \emph{non-official validation/testing sets} with one reference. Dataset$_1$ includes books overlapping with the training data, whereas dataset$_2$ features unseen books. Participants can treat each chapter as a document to train and test their discourse-aware models. Additional parallel training data from the General MT Task is also available for data augmentation. In the final test stage, systems translate the \emph{official testing set} (Test$_{final}$), which consists of approximately 20 consecutive chapters from each book, allowing participants to treat all chapters within a book as a long document\footnote{Participants may treat one chapter as a document, depending on model length capabilities.}. The final test set includes two references: Reference 1 by human translators and Reference 2 constructed through manual alignment of bilingual text on web pages. Genres in the validation and test sets are evenly sampled. for Dataset$_1$, books overlap with the training data, whereas Dataset$_2$ contains unseen books. Thus, each chapter is treated as a separate document. For Test$_{final}$, around 20 consecutive chapters from each book are selected, treating all chapters within a book as a long document. 

\item {\bf Zh-De and Zh-Ru sets}: Based on the GuoFeng Webnovel Corpus V1, we expand it from Chinese$\rightarrow$English to  Chinese$\rightarrow$German and Chinese$\rightarrow$Russian. Specifically, we translate Chinese web novels into German or Russian at the document level using GPT-4, and then have human translators review or post-edit the translations. The main difference from Zh-En set is that 1) there is no sentence-level alignment information in training/test sets; 2) there is only one reference in test sets; 3) there is only one official test set which is similar with Zh-En Test$_{final}$.

\end{itemize}

\subsection{Pretrained Models}
\label{sec:2-2}

In addition to the web novel training dataset, we provide in-domain pretrained models as supplementary resources (continuously trained on Chinese/English literary texts). These models can be used to fine-tune or initialize MT models. The RoBERTa (base) and mBART (CC25) models remain the same as last year, and this year we are providing an additional model, the Chinese-Llama-2 7B \cite{du-etal-2022-chinese-llama-2}.
Additionally, general-domain pretrained models listed in the General MT Track are permitted in this task: mBART, BERT, RoBERTa, sBERT, LaBSE.

\begin{table}[t]
\centering
% \scalebox{0.86}{
\begin{tabular}{p{1.8cm} p{5cm}}
\toprule
\bf Info. & {\bf Evaluator A}\\
\midrule
Position & Project manager and translator at a famous automobile enterprise \\
\cdashline{1-2}\noalign{\vskip 0.5ex}
Education & Master in English, a 211 comprehensive university\\
\cdashline{1-2}\noalign{\vskip 0.5ex}
Certification & TEM-8 \\
\cdashline{1-2}\noalign{\vskip 0.5ex}
Experience & Overseas project manager and translator for 10 years, part-time translator for 13 years \\
\midrule
\bf Level & {\bf Evaluator B}\\
\midrule
Position & Student at an international institution\\
\cdashline{1-2}\noalign{\vskip 0.5ex}
Education & Bachelor in French\\
\cdashline{1-2}\noalign{\vskip 0.5ex}
Certification & CET-6, IELTS-7.5\\
\cdashline{1-2}\noalign{\vskip 0.5ex}
Experience & Interpreter for international company and translation organization\\
\midrule

\bf Level & {\bf Evaluator C}\\
\midrule
Position & Translator at a translation company\\
\cdashline{1-2}\noalign{\vskip 0.5ex}
Education & Master of Marxism abroad\\
\cdashline{1-2}\noalign{\vskip 0.5ex}
Certification & CET-6\\
\cdashline{1-2}\noalign{\vskip 0.5ex}
Experience & Work in several translation companies\\
\bottomrule
\end{tabular}
\caption{The basic background of Zh-En human annotators.}
\label{tab:human-info}
\end{table}

\section{Evaluation Methods}
\label{sec:3}

The evaluation of document-level and literary-domain translation quality remains a complex challenge. In this year's shared task, the automated evaluation remains consistent with last year, and we are exploring new human evaluation methods, moving away from the multidimensional quality metrics (MQM) framework \cite{lommel2014multidimensional} used previously.

\subsection{Automatic Evaluation}

Our approach to automatic evaluation incorporates both sentence-level and document-level metrics. For sentence-level evaluation, we utilize sacreBLEU \cite{post2018call}, chrF \cite{popovic2015chrf}, TER \cite{snover2005study}, and the pretraining-based COMET \cite{rei2020comet}. For document-level evaluation, we employ document-level sacreBLEU (d-BLEU) \cite{liu2020multilingual}, which assesses n-gram matches across complete documents. This involves concatenating all sentences in a document into a single line before applying the sacreBLEU metric. Evaluations are conducted in a case-sensitive manner. We utilize the \texttt{sacrebleu} tool\footnote{\url{https://github.com/mjpost/sacrebleu} with signature: \texttt{nrefs:2|case:mixed|eff:no|tok:13a|smooth: exp|version:2.3.1}.} to calculate sacreBLEU, chrF, TER, and d-BLEU with two references. The command used is: \texttt{cat output | python -m sacrebleu reference*}. For COMET scores, we use the \texttt{unbabel-comet} tool\footnote{\url{https://github.com/Unbabel/COMET}.} with the command: \texttt{comet-score -s input -t output -r reference1} (default model).

\begin{table*}[t]
    \centering
    \scalebox{0.92}{
    \begin{tabular}{c c c c c c}
     \toprule
     \bf ID & {\bf Team} & {\bf Institution} & \bf Flag & \bf \# & \bf Main Methods\\
    \midrule
   1  & Cloudsheep & UC San Diego & $\bigotimes$ & 1 & \small Google Translate, GPT-4\\
   2  & HW-TSC & Huawei Translation Services Center & $\bigotimes$ & 2 &  \small Chinese-Llama2, Doc Decoding\\
   3  & NLP$^2$CT-UM & University of Macau & $\bigotimes$ & 5 & \small Qwen, GPT-4, DeepSeek\\
   4  & NTU & Nantong University & $\bigotimes$ & 1 & \small Finetuned Llama-3-Chinese, Phi-3\\
   5  & SJTU-LoveFiction & Shanghai Jiao Tong Universityy & $\bigotimes$ & 5 & \small Chunk-based SFT, Multi-Agents\\
    \bottomrule
    \end{tabular}}
    \caption{The summary of system submission and their participant teams. We also report the number of systems (\#) and the constrained ($\bigodot$) and unconstrained ($\bigotimes$) flags. }
    \label{tab:sys-summary}
\end{table*}

\subsection{Human Evaluation}

\paragraph{Guidelines}

We establish two sets of evaluation criteria: 1) {\em general quality}, covering aspects such as fluency and adequacy; 2) {\em discourse-aware quality}, including factors such as consistency, word choice, and anaphora. The detailed scoring criteria are listed in Table~\ref{tab:human_eval}. Accordingly, each output will be assigned two distinct scores (0$\sim$5). A score of 5 indicates excellent overall translation quality, with no grammatical errors, accurate word choice, consistent key terms, and consistent context and tone throughout the passage. A score of 0 indicates poor overall translation quality, with more than half of the translation being mistranslated or missing, inconsistent key terms, and poor fluency and clarity. In between scores reflect varying degrees of translation quality, with factors such as fluency, accuracy, consistency of key terms, and context and tone consistency affecting the score.
For each chapter, we assessed xxx sentences, with each instance containing outputs from 5 different systems. The scores were assigned to each window of neighboring sentences, taking into account the context provided by the entire document. Our intent was for evaluators to consider discourse properties beyond single sentences, while also avoiding the difficult task of evaluating an entire document. 
We employed two professional evaluators for our study. 

\paragraph{Human Evaluators}

We employed three professional evaluators and their background are detailed in Table~\ref{tab:human-info}. The human evaluations are only conducted for the Zh-En direction due to the limited participation in the Zh-De and Zh-Ru, with only two teams competing in each.
Besides, our annotators were given practice items, and the annotations reaches 0.86 Cohen's kappa scores \cite{mchugh2012interrater}, demonstrating that the annotators work efficiently and consistently under this guideline.

\section{Task Description}
\label{sec:4}

\paragraph{Overview}

The shared task focuses on translating literary texts across three language pairs: Chinese$\rightarrow$English, Chinese$\rightarrow$German, and Chinese$\rightarrow$Russian. Participants are provided with two types of training datasets: (1) the discourse-level GuoFeng Webnovel Corpus V2, with Chinese-English datasets offering sentence-level alignment, and new Chinese-German and Chinese-Russian datasets without alignment information; (2) General MT Track Parallel Training Data for broader translation tasks. Additionally, participants have access to pretrained models, including in-domain models like Chinese-Llama-2, In-domain RoBERTa (base), and In-domain mBART (CC25), as well as general-domain language models. 

In the final testing stage, participants will translate an official test set. The quality of translations is assessed using a combination of manual evaluation and automatic metrics. Systems are ranked based on human judgments according to our professional guidelines. Submissions are categorized into constrained or unconstrained tracks, depending on the data and resources used, with appropriate flags distinguishing them.

\paragraph{Goals}

The primary objectives of the task are to:
\begin{itemize}[leftmargin=*,topsep=0.1em,itemsep=0.1em,parsep=0.1em]
    \item Stimulate research in machine translation, focusing on literary texts and integrating discourse knowledge.
    \item Provide a platform for evaluating and comparing different methods and systems on a challenging dataset.
    \item Advance the state of the art in machine translation, with an emphasis on practical application scenarios.
\end{itemize}

\section{Participants' and Baseline Systems}
\label{sec:participants}

Here we briefly introduce each participant’s systems and refer the reader to the participant’s reports
for further details. Table \ref{tab:sys-summary} shows the summary of systems and participant teams.

\begin{table*}[t]
    \centering
    \scalebox{0.94}{
    \begin{tabular}{cc rrrrr}
     \toprule
     \multirow{2}{*}{\bf Type} & \multirow{2}{*}{\bf System} &  \multicolumn{4}{c}{\color{blue} Sent-Level}  & \multicolumn{1}{c}{\color{red} Doc-Level} \\
     \cmidrule(lr){3-6} \cmidrule(lr){7-7} 
     & & \bf BLEU$^\uparrow$ & \bf chrF2$^\uparrow$ & \bf COMET$^\uparrow$ & \bf TER$^\downarrow$ & \bf d-BLEU$^\uparrow$ \\
    \midrule
   \multirow{3}{*}{\em Baselines} & Google$^\star$ & 37.4 & 57.0 & 80.50 & 57.4 & 47.3\\
   & Llama-MT$^\star$ & n/a & n/a & n/a &  n/a& 43.1\\
    & GPT-4$^\star$ & n/a & n/a & n/a & n/a & 43.7\\
    % & TransAgents$^\star$ & n/a & n/a & n/a & n/a & --\\
    \midrule
    \cmidrule{1-7}
    \multirow{5}{*}{\em Primary} & Cloudsheep$^\star$ & 39.5 & 57.5 & 81.22 & 55.5 & 48.5 \\
    & HW-TSC$^\star$ & \bf {40.5} & \bf {58.5} & \bf {82.61} & \bf {56.0} & \bf {50.2}\\
    & NLP$^2$CT-UM$^\star$ & \bf {\color{blue}41.6} & \bf{\color{blue}58.7} & \bf{\color{blue}83.56} & \bf{\color{blue}52.7} & \bf{\color{red}50.9} \\
    & NTU$^\star$  & 20.9 & 41.9 & 74.53 & 73.9 & 34.6 \\
    & SJTU-LoveFiction$^\star$ & 35.1 & 54.7 & 80.79 & 62.1 & 47.2 \\
    \midrule
    \multirow{5}{*}{\em Contrastive} & HW-TSC$^\star$ & 40.6 & 58.6 & 82.59 & 55.9 & 50.3\\
    \cdashline{2-7}\noalign{\vskip 0.5ex}
    & NLP2CT-UM$_{1}^\star$ & 41.6 & 58.7 & 83.54 & 52.8 & 50.8 \\
    & NLP2CT-UM$_{2}^\star$ & 41.5 & 58.6 & 83.38 & 52.8 & 50.7 \\
    \cdashline{2-7}\noalign{\vskip 0.5ex}
    & SJTU-LoveFiction$_{1}^\star$ & 35.7 & 56.0 & 82.67 & 59.7 & 46.3 \\
    & SJTU-LoveFiction$_{2}^\star$ & 38.6 & 56.5 & 82.49 & 57.1 & 49.6 \\
    \bottomrule
    \end{tabular}}
    \caption{The Zh-En evaluation results of baseline and participants' systems in terms of {\bf automatic evaluation methods}, including 1) {\color{blue}sentence-level} metrics BLEU, chrF, COMET, TER; and 2) {\color{red}document-level} metrics d-BLEU. Systems marked with $^\star$ are unconstrained. The COMET is calculated with {\em unbabel-comet} using {\em Reference 1} while others are calculated with {\em sacrebleu} using two references. The ranked 1st and 2nd primary systems are highlighted.}
    \label{tab:auto-res}
\end{table*}

\subsection{Cloudsheep}
The team from UC San Diego, Halıcıoğlu Data Science Institute introduced 1 translation systems \cite{liu2024cloudsheep}. The CloudSheep system utilized a comprehensive tool pipeline to ensure translation accuracy and text coherence, especially emphasizing the consistent translation of names and idioms. Key strategies included the creation of custom dictionaries for names and the use of CEDICT for idiomatic expressions, with support from AI models like GPT-3.5-turbo and GPT-4 to refine translations. 

\subsection{HW-TSC}
The team from Huawei Translation Service Center introduced 2 translation systems \cite{luo2024context}. About the HW-TSC system, they enhanced the Chinese-Llama2 model with continual pre-training and SFT, incorporating an incremental decoding framework that considered broader textual contexts. This approach maintained narrative coherence and captured stylistic elements effectively, leading to marked improvements in BLEU scores at both the sentence and document levels.

\subsection{NLP$^2$CT-UM}
The team from NLP$^2$CT Lab, Department of Computer and Information Science, University of Macau introduced 5 translation systems \cite{liu2024noveltrans}.
Their approach utilized GPT-4o to generate three different translations using varied settings of additional contextual information and a custom terminology table. To optimize translation quality, they selected sentences with the highest xCOMET scores from among the alternatives for their final output, demonstrating a significant improvement over traditional chapter-level translation methods.

\subsection{NTU}

The team from Jiangsu Linchance Technology Co., Ltd. and Nantong University introduced 1 translation systems \cite{li2024linchance}. They leveraged the capabilities of the Llama and Phi models, employing both LoRA (Low-Rank Adaptation) and full-parameter tuning techniques. Despite memory constraints preventing the successful full-parameter tuning of the Llama-3-Chinese-8B-Instruct model, the fully fine-tuned Phi 3 model was chosen for its more natural and fluent output. Additionally, the integration of LoRA and a prompt-based translation system significantly enhanced the performance of the Llama3 model, outperforming other models in both BLEU and ROUGE metrics. 

\subsection{SJTU-LoveFiction}

The Shanghai Jiao Tong University (SJTU LoveFiction) team implemented advanced methodologies across Chinese to English, German, and Russian pairs \cite{sun-etal-2024-final}. The SJTU-LoveFiction system integrated the Qwen2-72B, Claude3.5, and GPT-4o models and featured novel techniques such as chunk-based SFT for enhanced contextual coherence, a multi-model merging approach using a GPT-4o-based Translation Editor Agent to synthesize translations, and a Terminology Intervention process employing a Term Proofreader Agent to ensure terminology consistency.

\section{Evaluation Results}
\label{sec:evaluation}

\subsection{Automatic Evaluation}

\paragraph{Zh-En.}

Table~\ref{tab:auto-res} demonstrates a comprehensive overview of the performance of various translation systems across the Chinese to English language pair. The evaluation was based on both sentence-level and document-level automatic metrics, including BLEU, chrF2, COMET, TER, and d-BLEU. 

At the sentence level, NLP$^2$CT-UM achieved the highest scores in BLEU (41.6) and COMET (83.56), indicating superior translation quality and contextual alignment with the reference text. The chrF2 score, which assesses character n-gram precision and recall, was also notably high for NLP$^2$CT-UM at 58.7. Besides, NLP$^2$CT-UM achieves the lowest score in TER (52.7), suggesting it had fewer translation errors relative to the length of the output. At the document level, the d-BLEU metric, which evaluates the coherence and flow of translations over entire documents, highlighted NLP$^2$CT-UM as a standout performer with a score of 50.9, slightly above HW-TSC with 50.2 and significantly ahead of other systems.

At the document level, the primary systems outperform the baselines in terms of d-BLEU, reflecting better overall translation consistency. Among the baselines, Google achieves 47.3, while the primary system NLP$^2$CT-UM surpasses it with a score of 50.9. Similarly, HW-TSC achieves a strong d-BLEU of 50.2, showing consistent improvements over the baseline. These results highlight the primary systems’ ability to produce more accurate and coherent translations over entire documents compared to the baselines.

\begin{table}[t]
    \centering
    \scalebox{0.94}{
    \begin{tabular}{ccc}
     \toprule
     \multirow{2}{*}{\bf System} & \multicolumn{2}{c}{\color{red} d-BLEU$^\uparrow$} \\
     \cmidrule(lr){2-3}
     & Zh-De & Zh-Ru \\
    \midrule
   Google$^\star$ & \bf 31.3 & \bf 25.2 \\
    GPT-4$^\star$ & 26.7 & 20.5 \\
    \midrule
    NLP$^2$CT-UM$^\star$ & \bf 26.7 & \bf 22.7\\
    SJTU-LoveFiction$^\star$ & 25.4 & 21.5\\
    \bottomrule
    \end{tabular}}
    \caption{The Zh-De and Zh-Ru evaluation results of baseline and participants' systems in terms of {\bf automatic evaluation methods (d-BLEU)}. Systems marked with $^\star$ are unconstrained. The best primary systems are highlighted.}
    \label{tab:auto-res-deru}
\end{table}

\paragraph{Zh-De and Zh-Ru.}

Table \ref{tab:auto-res-deru} demonstrates the results on Zh-De and Zh-Ru directions. The Google baseline achieves the highest d-BLEU scores of 31.3 and 25.2, respectively, demonstrating its strong overall translation performance for these language pairs. Among the primary systems, NLP$^2$CT-UM achieves 26.7 in Zh-De and 22.7 in Zh-Ru, while SJTU-LoveFiction scores slightly lower with 25.4 and 21.5. These results indicate that while the primary systems show competitive performance, they fall short of the Google baseline, especially in maintaining document-level consistency for these language pairs.

\begin{table}[t]
    \centering
    \scalebox{0.95}{
    \begin{tabular}{c r c c}
     \toprule
     {\bf System} & {\color{blue} General} & {\color{red}Discourse} &  {\color{forestgreen} Rank}\\
    \midrule
    Cloudsheep & 3.67 & 3.71 & 3\\
    HW-TSC & \bf 3.54  & \bf 3.46 & 4\\
    NLP2CT-UM & {\color{blue} \bf 3.96}  & 4.00 & 1 / 2\\
    NTU & 2.58  & 2.42 & 5\\
    SJTU-LoveFiction & 3.92  & {\color{red} \bf 4.13} & 2 / 1\\
    \bottomrule
    \end{tabular}}
    \caption{The Zh-En evaluation results of baseline and primary systems in terms of {\bf human evaluation}. We report two types of scores and System Rank. The human evaluation based on a scale from 0-5 encompasses two dimensions: general quality and discourse awareness.}
    \label{tab:human-res}
\end{table}

\subsection{Human Evaluation}

In the Zh-En human evaluation, the systems were assessed based on general quality and discourse awareness, both scored on a scale of 0-5.
\begin{itemize}[leftmargin=*,topsep=0.1em,itemsep=0.1em,parsep=0.1em]

\item {\em NLP$^2$CT-UM} achieved the highest scores in General (3.96) and Discourse (4.00), securing the top ranks (1st in General, tied 2nd in Discourse), indicating strong overall translation quality and contextual coherence.

\item {\em SJTU-LoveFiction} performed exceptionally well in Discourse with a score of 4.13, the highest in this category, and scored 3.92 in General, earning 2nd place in General and 1st place in Discourse.

\item {\em Cloudsheep} achieved balanced scores of 3.67 (General) and 3.71 (Discourse), ranking 3rd overall.

\item {\em HW-TSC} scored 3.54 (General) and 3.46 (Discourse), placing 4th in both categories.

\item {\em NTU} lagged behind with 2.58 (General) and 2.42 (Discourse), ranking 5th overall.

\end{itemize}

These results highlight NLP$^2$CT-UM and SJTU-LoveFiction as the top-performing systems, excelling in both general translation quality and discourse-level awareness.

\section{Conclusion}

Building on our experience in discourse-aware machine translation, in the era of NMT and \cite{wang2017exploiting, lyu2023new, wang2020tencent, wang2019one, longyue2019discourse} LLM \cite{pang2024salute, lyu2023paradigm, ming2024marco, wu2024transagents}, we continue to organize the WMT 2024 Shared Task: Discourse-Level Literary Translation, reporting the latest methods and advancements for this challenge.

\bibliography{anthology,custom}
\bibliographystyle{acl_natbib}

% \appendix

% \section{Example Appendix}
% \label{sec:appendix}

% This is a section in the appendix.

\begin{table*}[t]
    \centering
    \scalebox{0.95}{
    \begin{tabular}{c p{6.8cm} p{6.8cm}}
    \toprule
        \textbf{Score} & \textbf{General Quality} & \textbf{Discourse Awareness}\\
    \midrule
    5 & Translation passes quality control; the overall translation is excellent. 
Translation is very fluent with no grammatical errors and has been localized to fit target language. Word choice is accurate with no mistranslations. The translation is a 100\% true to the source text. & No inconsistency relating to key terms such as names, organization, etc. Linking words or expressions between sentences keeps the logic and language of the passage clear and fluent. Context and tone are consistent throughout. The style of the text conforms to the culture and habit of the target language. \\
\midrule
4 & Translation passes quality control; the overall translation is very good. Translation is fluent. Any errors that may be present does not affect the meaning or comprehension of the text. Most word choice is accurate, but some may cause ambiguity. 	Key terms are consistent. Inconsistency is limited to non-key terms. & Logical and language is clear and fluent. Some sentences lack transition but does not affect contextual comprehension. Topic is consistent. Tone and word choice may be inconsistent, but comprehension is not affected. Translation conforms to the culture and habit.\\
\midrule
3 & Translation passes quality control; the overall translation is ok. 
Translation is mostly fluent but there are many sections that require rereading due to language usage. Some word choice is inaccurate or errors but meaning of the sentence can be inferred from context. &	Some key terms may be inconsistent. Most sentences translation smoothly and logically but some sentences that may seem abrupt due to lack of linkage. 
Topic is consistent. Tone and word choice is inconsistent, noticeably affecting the accuracy of reading comprehension. \\
\midrule
2 & Translation does not pass quality control; the overall translation is poor. Meaning is unclear or disjointed. Even with multiple rereading, passage may still be incomprehensible. Translation is not accurate to the source text or is missing in large quantities, causing the translation to deviate from the source text. & Many key terms are inconsistent, needing multiple rereading to understand context of the passage. Some linkages are present but overall, the passage lacks fluency and clarity, causing trouble with comprehension. The topic or tone is different from the other passages, affecting reading comprehension. \\
\midrule
1 & Translation does not pass quality control; the overall translation is very poor. More than half of the translation is mistranslated or missing. & Key terms are inconsistent, causing great trouble with comprehension. Some linkages are present but overall, the passage lacks fluency and clarity, heavily interfering with comprehension. The topic or tone is different from the other passages, heavily interfering with comprehension. \\
\midrule
0 & Translation output is unrelated to the source text. & Output is unrelated to previous or following sections. \\
    \bottomrule
    \end{tabular}}
    \caption{Human evaluation criteria on document-level translation.}
    \label{tab:human_eval}
\end{table*}

\end{document}